\documentclass{article}
\pdfoutput=1

     \usepackage[final]{neurips_2019}

\usepackage[utf8]{inputenc} 
\usepackage[T1]{fontenc}    
\usepackage{hyperref}       
\usepackage{url}            
\usepackage{booktabs}       
\usepackage{amsfonts}       
\usepackage{nicefrac}       
\usepackage{microtype}      
\usepackage{graphicx}
\usepackage{makecell}
\usepackage{xcolor}
\usepackage{subfig}

\title{Hidden Stratification Causes Clinically Meaningful Failures in Machine Learning for Medical Imaging
}

\author{%
  Luke Oakden-Rayner,\thanks{Equal Contribution} $ $ Gustavo Carneiro\\
  Australian Institute for Machine Learning\\
  University of Adelaide\\
  Adelaide, SA 5000 \\
  \texttt{\{luke.oakden-rayner,gustavo.carneiro\}} \\
  \texttt{@adelaide.edu.au} \\
  \And
    Jared Dunnmon,$^*$ Christopher R\'{e}\\
  Department of Computer Science\\
  Stanford University\\
  Stanford, CA 94305 \\
  \texttt{\{jdunnmon,chrismre\}} \\
  \texttt{@stanford.edu} 
  }

\usepackage[compact]{titlesec}         
\titlespacing{\section}{0pt}{0pt}{0pt} 
\AtBeginDocument{
  \setlength\abovedisplayskip{0pt}
  \setlength\belowdisplayskip{0pt}}
  
\begin{document}

\maketitle


\vspace{-5 mm}
\section{Introduction}

Deep learning systems have shown remarkable promise in medical image analysis, often claiming performance rivaling that of human experts \citep{esteva2019guide}. 
 However, performance results reported in the literature may overstate the clinical utility and safety of these models.  
 Specifically, it is well known that machine learning models often make mistakes that humans never would, despite having aggregate error rates comparable to or better than those of human experts. An example of this ``inhuman'' lack of common sense might include a high performance system that calls any canine in the snow a wolf, and one on grass a dog, regardless of appearance \citep{ribeiro2016should}.
While this property of machine learning models has been underreported in non-medical tasks---possibly because safety is often less of a concern and all errors are roughly equivalent in cost---it likely to be of critical importance in medical practice, where specific types of errors can have serious clinical impacts. 
 
Of particular concern is the fact that most medical machine learning models are built and tested using an incomplete set of possible labels---or \textit{schema}---and that the training labels therefore only coarsely describe the meaningful variation within the population. 
Medical images contain dense visual information, and imaging diagnoses are usually identified by recognizing the combination of several different visual features or patterns. 
This means that any given pathology or variant defined as a ``class'' for machine learning purposes is often comprised of several visually and clinically distinct subsets; a ``lung cancer'' label, for example, would contain both solid and subsolid tumours, as well as central and peripheral neoplasms. 
We call this phenomenon \textit{hidden stratification}, meaning that the data contains unrecognized subsets of cases which may affect model training, model performance, and most importantly the clinical outcomes related to the use of a medical image analysis system.  

Worryingly, when these subsets are not labelled, even performance measurements on a held-out test set may be falsely reassuring. 
This is because aggregate performance measures such as accuracy, sensitivity (i.e.~recall), or ROC AUC can be dominated by larger subsets, obscuring the fact that there may be an unidentified subset of cases within which performance is poor. 
Given the rough medical truism that serious diseases are less common than mild diseases, it is even likely that underperformance in minority subsets could lead to disproportionate harm to patients.

We describe three different techniques for measuring hidden stratification effects -- schema completion, error auditing, and algorithmic measurement -- and use them to show not only that hidden stratification can result in performance differences of up to 20\% on clinically important subsets, but also that simple unsupervised learning approaches can help to identify these effects.  
Across datasets, we find evidence that hidden stratification occurs on subsets characterized by a combination of low prevalence, poor label quality, subtle discriminative features, and spurious correlates. 
We examine the clinical implications of these findings, and argue that measurement and reporting of hidden stratification effects should become a critical component of machine learning deployments in medicine.

\section{Methods for Measuring Hidden Stratification}
We examine three possible approaches to measure the clinical risk of hidden stratification: (1) exhaustive prospective human labeling of the data, called \textit{schema completion}, (2) retrospective human analysis of model predictions, called \textit{error auditing}, and (3) \textit{algorithmic methods} to detect hidden strata.  
Each method can be applied to the test dataset, allowing for analysis and reporting (e.g., for regulatory processes) of subclass (i.e.~subset) performance without re-labeling large training sets.

\textbf{Schema Completion}: In schema completion, the schema author prospectively prescribes a more complete set of subclasses that need to be labeled, and provides these labels on test data. 
Schema completion has many advantages, such as the ability to prospectively arrive at consensus on subclass definitions (e.g.~a professional body could produce standards describing reporting expectations) to both enable accurate reporting and guide model development.
However, schema completion is fundamentally limited by the understanding of the schema author; if important subclasses are omitted, schema completion does not protect against important clinical failures.
Further, it can be time consuming (or practically impossible!) to exhaustively label all possible subclasses, which in a clinical setting might include subsets of varying diagnostic, demographic, clinical, and descriptive characteristics.
Finally, a variety of factors including the visual artifacts of new treatments and previously unseen pathologies can render existing schema obsolete at any time.

\textbf{Error Auditing}: In error auditing, the auditor examines model outputs for unexpected regularities such as a consistently incorrect model prediction on a recognizable subclass. 
Advantages of error auditing include that it is not limited by predefined expectations of schema authors, and that the space of subclasses considered is informed by model function.
Rather than having to enumerate every possible subclass, only subclasses observed to be concerning need be measured.
While more labor-efficient than schema completion, error auditing is critically dependent on the ability of the auditor to visually recognize anomalous patterns in the distribution of model outputs.
It is therefore more likely that the non-exhaustive nature of audit could limit certainty that all important strata were analyzed.
Of particular concern is the ability of error auditing to identify low-prevalence, high discordance subclasses that may rarely occur but are clinically salient.

\textbf{Algorithmic Measurement}: In algorithmic measurement approaches, the algorithm developer designs a method to search for subclasses automatically. 
In many cases, such algorithms will be unsupervised methods such as clustering. 
If any identified group (e.g.~a cluster) underperforms compared to the overall superclass, then this may indicate the presence of a clinically relevant subclass.
Clearly, the use of algorithmic approaches still requires human review in a manner that is similar to error auditing, but is less dependent on the specific human auditor to initially identify the stratification.  
While algorithmic approaches to measurement can reduce burden on human analysts and take advantage of learned encodings to identify subclasses, their efficacy is limited by the difficulty of separating important subclasses in the feature space analyzed.

\section{Experiments}

We empirically measure the effect of hidden stratification in medical imaging using each of these approaches.  
We hypothesize that there are several subclass characteristics that contribute to degraded model performance: (1) low subclass prevalence, (2) reduced label accuracy within the subclass, (3) subtle discriminative features, and (4) spurious correlations \citep{Selbst2017-gz}. 
We first use schema completion to evaluate clinically important hidden stratification effects in radiograph datasets describing hip fracture (low subclass prevalence, subtle discriminative features) and musculoskeletal extremity abnormalities (poor label quality, subtle discriminative features).
We then demonstrate how error auditing can be used to identify hidden stratification in a large public chest radiograph dataset that contains a spurious correlate.
Finally, we show that a simple unsupervised clustering algorithm can provide value in some cases by approximately separating the well-performing and poorly-performing subclasses .
\vspace{- 4 mm}
\paragraph{Schema Completion:} 
Schema completion indicates the presence of hidden stratification on a large, high quality pelvic x-ray dataset from the Royal Adelaide Hospital \citep{Gale_W_Oakden-Rayner_L_Carneiro_G_Bradley_AP_Palmer_LJ2017-tl}.
 A DenseNet model previously trained on this dataset to identify hip fractures achieved extremely high performance (ROC AUC = 0.994) \citep{Gale_W_Oakden-Rayner_L_Carneiro_G_Bradley_AP_Palmer_LJ2017-tl}. 
 The distribution of the location and description subclasses is shown in Table \ref{tab:hip1}, with subclass labels produced by a board-certified radiologist (LOR).  
 We find that sensitivity on both subtle fractures (0.900) and low-prevalence cervical fractures (0.911) is significantly lower (p $<$ 0.01) than that on the overall task (0.981).
These results support the hypothesis that both subtle discriminative features and low prevalence can contribute to clinically relevant stratification.  

\begin{figure}[htb!]%
 \vspace{-6mm}
\centering
\subfloat[]{\includegraphics[trim={0 0 1cm 1.5cm}, clip, width=2.5in]{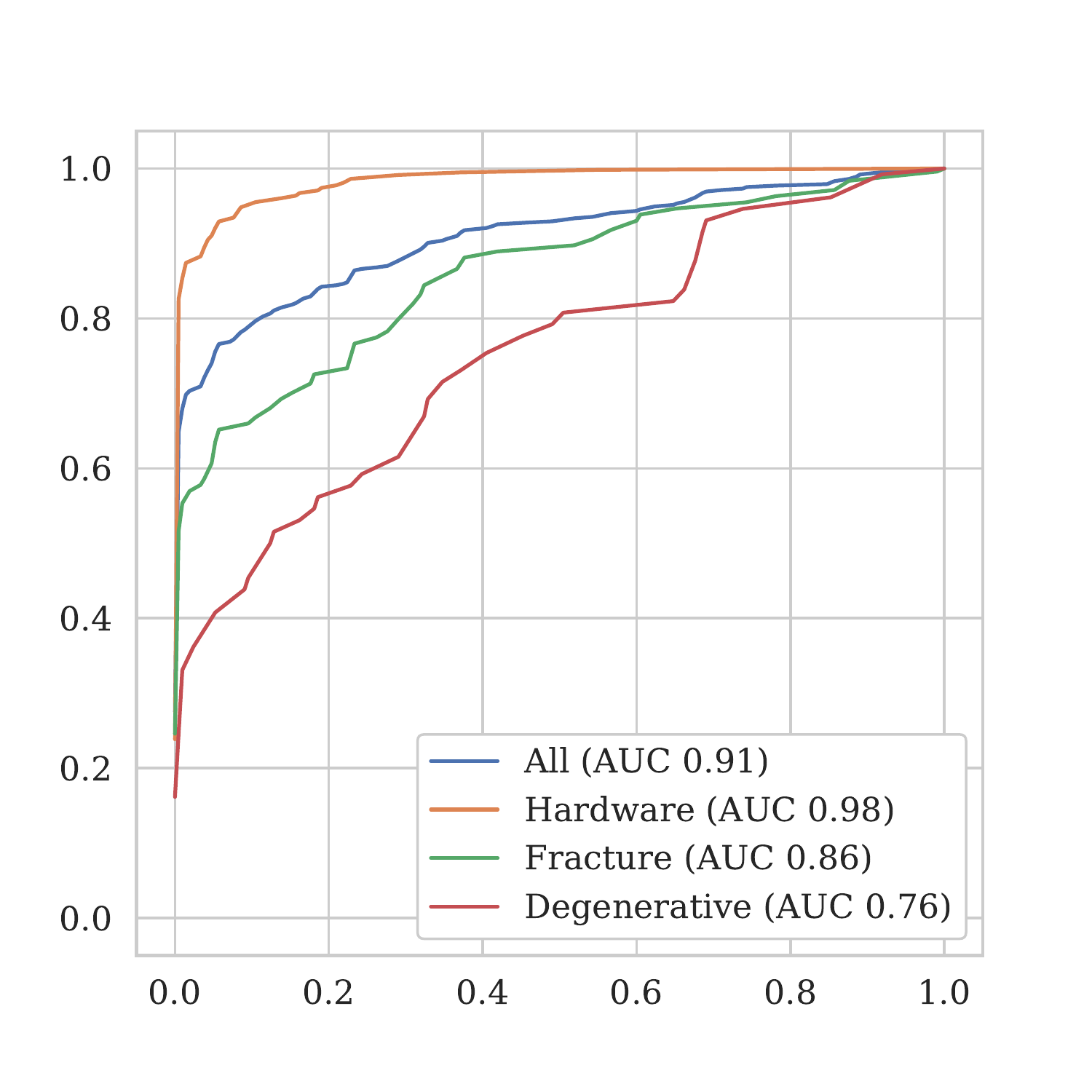}}%
\subfloat[]{\includegraphics[trim={0 0 1cm 1.5cm}, clip, width=2.5in]{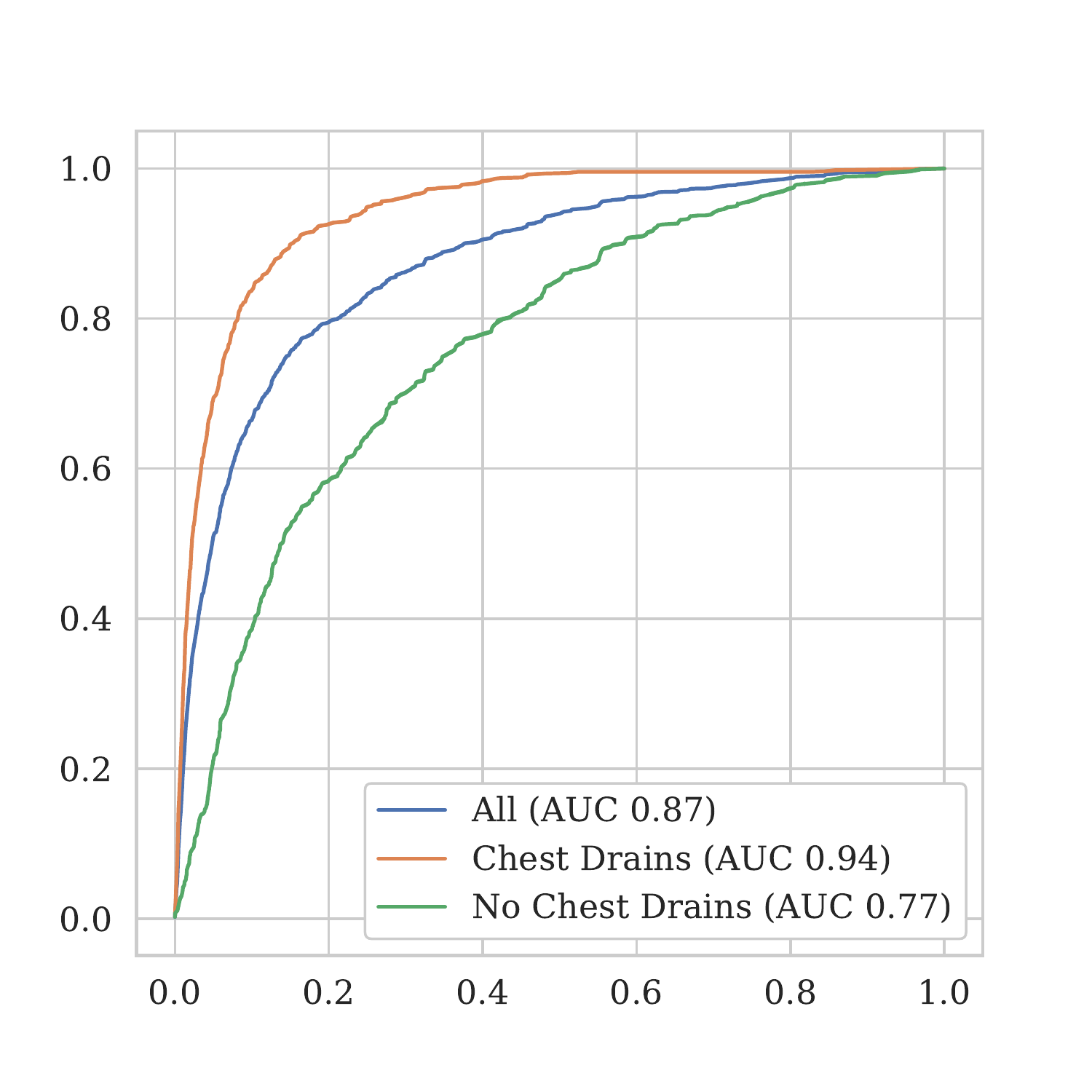}}
\caption{ROC curves for subclasses of the (a) abnormal MURA superclass and (b) pneumothorax CXR14 superclass. All subclass AUCs are significantly different than the overall task (p $<$ 0.05).}
\label{fig:rocs}
\vspace{-5 mm}
\end{figure}

We next use schema completion to demonstrate the effect of hidden stratification on the MURA musculoskeletal x-ray dataset developed by Rajpurkar et al. \citep{Rajpurkar2017-rc}, which provides labels for a single class that indicates whether each case is ``normal'' or ``abnormal.'' 
These binary labels have been previously investigated and relabelled with subclass identifiers by a board-certified radiologist \citep{Oakden-Rayner2019-yi}, showing substantial differences in both the prevalence and sensitivity of the labels within each subclass (see Table \ref{tab:mura2}). 
While this schema remains incomplete, even partial schema completion demonstrates substantial hidden stratification in this dataset.
 We train a DenseNet-169 on the normal/abnormal labels, with 13,942 cases used for training and 714 cases held-out for testing \citep{Rajpurkar2017-rc}.  
 In Fig.~\ref{fig:rocs}(a), we present ROC curves and ROC AUC values for each subclass and in aggregate.  
 We find that overall ROC AUC for the easy-to-detect subclass containing hardware (0.98) is higher than aggregate ROC AUC (0.91), despite the low subclass prevalence.
 As expected, we also observe degraded ROC AUC for degenerative disease (0.76), which has low-sensitivity superclass labels and subtle visual features (Table \ref{tab:mura2}).  

\paragraph{Error Auditing:} We next use error auditing to show that the clinical utility of a common model for classifying the CXR14 chest radiograph dataset \citep{Wang2017-vm} may be substantially reduced by hidden stratification effects in the pneumothorax class that result from spurious correlates.
This dataset contains 112,120 frontal chest films from 30,805 unique patients, and each image was labeled for 14 different thoracic pathologies. 
In our analysis, we leverage a pretrained Densenet-121 model provided by Zech \citep{Zech_undated-cw} which reproduces the procedure and results of Rajpurkar et al. \citep{Rajpurkar2018-gc} on this dataset.  

During error auditing, where examples of false positive and false negative predictions from the pretrained model were visually reviewed by a board certified radiologist \citep{Oakden-Rayner2019-yi},
it was observed that pneumothorax cases without chest drains were highly prevalent in the set of false negatives.
A chest drain is a non-causal image feature in the setting of pneumothorax, as this device is the common form of \textit{treatment} for the condition. 
As such, not only does this reflect a spurious correlate, but the correlation is in fact highly clinically relevant; untreated pneumothoraces are life-threatening while treated pneumothoraces are usually benign.
 To explore this audit-detected stratification, pneumothorax subclass labels for ``chest drain'' and ``no chest drain'' were provided by a board-certified radiologist (LOR) for each element of the test set.  
 Due to higher prevalence of scans with chest drains in the dataset, clear discriminative features of a chest drain, and high label quality for the scans with chest drains, we hypothesize that a model trained on the CXR14 dataset will attain higher performance on the pneumothorax subclass with chest drains than that without chest drains.  
 

We present ROC curves for each pneumothorax subclass in Fig.~\ref{fig:rocs}(b).  
While overall pneumothorax ROC AUC closely matches that reported in Rajpurkar et al.~\citep{rajpurkar2017chexnet} at 0.87, pneumothorax ROC AUC was 0.94 on the subclass with chest drains, but only 0.77 on the subclass without chest drains.  
We find that 80\% of pneumothoraces in the test set contained a chest drain, and that positive predictive value on this subclass was 30\% higher (0.90) than on those with no chest drain (0.60).  
These results suggest that clearly identifiable spurious correlates can also cause clinically important hidden stratification.

\paragraph{Algorithmic Measurement with Unsupervised Clustering:} While schema completion and error auditing have allowed us to identify hidden stratification problems in multiple medical machine learning datasets, each requires substantial effort from clinicians.
Further, in auditing there is no guarantee that an auditor will recognize underlying patterns in the model error profile.
In this context, unsupervised learning techniques can be valuable tools in automatically identifying hidden stratification.
We show that even simple k-means clustering can detect several of the hidden subclasses identified above via time-consuming human review or annotation.

For each superclass, we apply k-means clustering to the pre-softmax feature vector of all test set examples within that superclass using $k \in \{2,3,4,5\}$.
For each value of $k$, we select the two clusters with greater than 100 constituent points that have the largest difference in error rates (to select a ``high error cluster'' and ``low error cluster'' for each $k$).
Finally, we return the pair of high and low error clusters that have the largest Euclidean distance between their centroids.
Ideally, examining these high and low error clusters would help human analysts identify salient stratifications in the data.
Note that our clustering hyperparameters were coarsely tuned, and could likely be improved in practice.

To evaluate the utility of this approach, we apply it to several datasets analyzed above, and report results in Table \ref{tab:clustercifar-1}.  
We find that while this simple k-means clustering approach does not always yield meaningful separation (e.g.~on MURA), it does produce clusters with a high proportion of drains on CXR14. 
 In practice, such an approach could be used both to assist human auditors in identifying salient stratifications in the data and to confirm that schema completion has been successful.

 \begin{table}[]
 \centering
\begin{tabular}{ccc}
\toprule
 Dataset-Superclass (Subclass) & \makecell{Difference in Subclass Prevalence \\ (High Error Cluster, Low Error Cluster)}  & \makecell{Overall Subclass \\ Prevalence} \\
 \toprule
 CXR14-Pneumothorax (Drains) & 0.68 (0.17, 0.84) & 0.80\\
 MURA-Abnormal (Hardware) & 0.03 (0.29, 0.26) & 0.11\\
 MURA-Abnormal (Degenerative) & 0.04 (0.12, 0.08) & 0.43\\
 \toprule
\end{tabular}
\caption{ Subclass prevalence in high and low error clusters on CXR14 and MURA.}
\label{tab:clustercifar-1}
\vspace{- 10mm}
\end{table}

\section{Discussion}

We find evidence that hidden stratification can lead to markedly different superclass and subclass performance when labels for the subclasses have different levels of accuracy, when the subclasses are imbalanced, when discriminative visual features are subtle, or when spurious correlates such as chest drains are present.
The clinical implications of hidden stratification will vary by task. 
Our MURA results, for instance, are unlikely to be clinically relevant, because degenerative disease is rarely a significant or unexpected finding, nor are rapid complications likely. 
We hypothesize that labels derived from clinical practice are likely to demonstrate this phenomenon; that irrelevant or unimportant findings are often elided by radiologists, leading to reduced label quality for less significant findings.

The findings in the CXR14 task are far more concerning. 
The majority of x-rays in the pneumothorax class contain chest drains, the presence of which is a healthcare process variable that is not causally linked to pneumothorax diagnosis.
 Importantly, the presence of a chest drain means these pneumothorax cases are already treated and are therefore at much less risk of pneumothorax-related harm. 
 In this experiment, we see that the performance in the clinically important subclass of cases without chest drains is far worse than the overall task results would suggest. 
 We could easily imagine a similar situation where a model is incorrectly justified for clinical use or regulatory approval based on the results from the overall task alone; such a scenario could ultimately cause harm to patients.
 
While the CXR14 example is quite extreme, it does correspond with the medical truism that serious disease is typically less common than non-serious disease. 
These results suggest that image analysis systems that appear to perform well on a given task may actually fail to identify the most clinically important cases. 
This behavior is particularly concerning when comparing these systems to human experts, who focus a great deal of effort on specifically learning to identify rare, dangerous, and subtle disease variants.
Encouragingly, we do find evidence that a simple unsupervised approach to identify unrecognized subclasses can produce clusters containing different proportions of examples from the hidden subclasses our analysis had previously identified. 
In summary, the findings presented here highlight the largely unrecognized problem of hidden stratification in clinical imaging datasets, and suggest that awareness of hidden stratification is important and should be considered (even if to be later dismissed) when planning, building, evaluating, and regulating clinical image analysis systems.
 



\bibliographystyle{plain}
\bibliography{mi-subsets.bib}

\begin{thebibliography}{10}

\bibitem{esteva2019guide}
Andre Esteva, Alexandre Robicquet, Bharath Ramsundar, Volodymyr Kuleshov, Mark
  DePristo, Katherine Chou, Claire Cui, Greg Corrado, Sebastian Thrun, and Jeff
  Dean.
\newblock A guide to deep learning in healthcare.
\newblock {\em Nature Medicine}, 25(1):24, 2019.

\bibitem{Gale_W_Oakden-Rayner_L_Carneiro_G_Bradley_AP_Palmer_LJ2017-tl}
William Gale, Luke Oakden-Rayner, Gustavo Carneiro, Andrew Bradley, and Lyle
  Palmer.
\newblock Detecting hip fractures with radiologist-level performance using deep
  neural networks.
\newblock {\em arXiv preprint arXiv:1711.06504}, 2017.

\bibitem{Oakden-Rayner2019-yi}
Luke Oakden-Rayner.
\newblock Exploring large scale public medical image datasets.
\newblock {\em arXiv preprint arXiv:1907.12720}, July 2019.

\bibitem{Rajpurkar2017-rc}
Pranav Rajpurkar, Jeremy Irvin, Aarti Bagul, Daisy Ding, Tony Duan, Hershel
  Mehta, Brandon Yang, Kaylie Zhu, Dillon Laird, Robyn~L Ball, et~al.
\newblock {MURA}: Large dataset for abnormality detection in musculoskeletal
  radiographs.
\newblock {\em arXiv preprint arXiv:1712.06957}, 2017.

\bibitem{Rajpurkar2018-gc}
Pranav Rajpurkar, Jeremy Irvin, Robyn~L Ball, Kaylie Zhu, Brandon Yang, Hershel
  Mehta, Tony Duan, Daisy Ding, Aarti Bagul, Curtis~P Langlotz, Bhavik~N Patel,
  Kristen~W Yeom, Katie Shpanskaya, Francis~G Blankenberg, Jayne Seekins,
  Timothy~J Amrhein, David~A Mong, Safwan~S Halabi, Evan~J Zucker, Andrew~Y Ng,
  and Matthew~P Lungren.
\newblock Deep learning for chest radiograph diagnosis: A retrospective
  comparison of the {CheXNeXt} algorithm to practicing radiologists.
\newblock {\em PLoS Medicine}, 15(11):e1002686, November 2018.

\bibitem{rajpurkar2017chexnet}
Pranav Rajpurkar, Jeremy Irvin, Kaylie Zhu, Brandon Yang, Hershel Mehta, Tony
  Duan, Daisy Ding, Aarti Bagul, Curtis Langlotz, Katie Shpanskaya, Matthew
  Lungren, and Andrew Ng.
\newblock Chexnet: Radiologist-level pneumonia detection on chest x-rays with
  deep learning.
\newblock {\em arXiv preprint arXiv:1711.05225}, 2017.

\bibitem{ribeiro2016should}
Marco~Tulio Ribeiro, Sameer Singh, and Carlos Guestrin.
\newblock Why should {I} trust you?: Explaining the predictions of any
  classifier.
\newblock In {\em Proceedings of the 22nd ACM SIGKDD International Conference
  on Knowledge Discovery and Data Mining}, pages 1135--1144. ACM, 2016.

\bibitem{Selbst2017-gz}
Andrew~D Selbst.
\newblock Disparate impact in big data policing.
\newblock {\em Georgia Law Review}, 52:109, 2017.

\bibitem{Wang2017-vm}
Xiaosong Wang, Yifan Peng, Le~Lu, Zhiyong Lu, Mohammadhadi Bagheri, and
  Ronald~M Summers.
\newblock Chest x-ray 8: Hospital-scale chest x-ray database and benchmarks on
  weakly-supervised classification and localization of common thorax diseases.
\newblock In {\em {IEEE} Conference on Computer Vision and Pattern Recognition
  ({CVPR)}}, pages 3462--3471, 2017.

\bibitem{Zech_undated-cw}
John Zech.
\newblock reproduce-chexnet, 2019.

\end{thebibliography}

\clearpage
\section*{Appendix}

\begin{table}[htb!]
\vspace{-2mm}
\centering
\begin{tabular}{ccc}
\toprule
 Subclass & Prevalence (Count) & Sensitivity \\
 \toprule
 Overall & 1.00 (643) & 0.981  \\
 Subcapital & 0.26 (169) & 0.987   \\
 Cervical & 0.13 (81) & \textbf{0.911}\\
 Pertrochanteric & 0.50 (319)  & 0.997\\
 Subtrochanteric & 0.05 (29) & 0.957 \\
 Subtle & 0.06 (38) & \textbf{0.900}\\
 Mildly Displaced & 0.29 (185) & 0.983\\
 Moderately Displaced & 0.30 (192) & 1.000\\
 Severely Displaced & 0.36 (228) & 0.996\\
 Comminuted & 0.26 (169) & 1.000 \\ 
 \toprule
\end{tabular}
\caption{Superclass and subclass performance for hip fracture detection from frontal pelvic x-rays. Bolded subclasses show significantly worse performance than that on the overall task.}
\label{tab:hip1}
\vspace{-6mm}
\end{table}

\begin{table}[!h!]
\centering
\begin{tabular}{ccc}
 \toprule
 Subclass & Subclass Prevalence & Superclass Label Sensitivity \\
 \toprule
 Fracture & 0.30 & 0.92   \\
 Hardware & 0.11 & 0.85    \\
 DJD & 0.43 & 0.60 \\
 \toprule
\end{tabular}
\caption{MURA ``abnormal'' label prevalence and sensitivity for the subclasses of ``fracture,'' ``hardware,'' and ``degenerative joint disease (DJD).'' The degenerative joint disease subclass labels have the highest prevalence but the lowest sensitivity with respect to review by a board-certified radiologist.}
\label{tab:mura2}
\vspace{-8mm}
\end{table}

\end{document}